%% file: Main.tex
\documentclass[conference]{IEEEtran}

\usepackage{graphicx}
\usepackage{amsmath}
\usepackage{amssymb}
\usepackage{algorithm}
\usepackage{algorithmicx}
\usepackage{algpseudocode}
\usepackage{amsfonts,amssymb,amsthm}
\usepackage{svn-multi}
\usepackage{etoolbox}
\let\labelindent\relax
\usepackage{breqn}
\usepackage{subcaption}
\usepackage{enumitem}
\setlist{
	topsep=0pt,
	partopsep=0pt,
	itemsep=0pt,
	parsep=0pt,
	leftmargin=*,
	nolistsep
}

\makeatletter
\patchcmd{\@makecaption}
{\scshape}
{}
{}
{}
\makeatother
\usepackage{caption}
\newcount\outlineon
\newcommand{\outline}[1]{\ifnum\outlineon>0\
	\\\noindent\fbox{\begin{minipage}{\linewidth}{\footnotesize#1}\end{minipage}}\\\\\fi}
\newcommand{\execlude}[1]{}

\newcommand{\proofend}[2]{ $ \blacksquare$ \vspace{5mm} }

\theoremstyle{plain}

\theoremstyle{definition}

\theoremstyle{remark}

\newif\ifcomments

\commentstrue 
\commentsfalse 

\input{comments}

\begin{document}
	
	\ifcomments
	\outlineon = 1
	\fi
	
	%
	\title{Safety-Critical Edge Robotics Architecture with Bounded End-to-End Latency}
	

\author{
	\IEEEauthorblockN{Gautam Gala, Luiz Maia, Isser Kadusale,\\Mohammad Ibrahim Alkoudsi and Gerhard Fohler}
	\IEEEauthorblockA{
		Chair of Real-Time Systems,\\
		Technical University of Kaiserslautern-Landau (RPTU)\\
		\{gala,maia,kadusale,alkoudsi,fohler\}@eit.uni-kl.de
	}
	\and
	\IEEEauthorblockN{Tilmann Unte, Johannes Kühbacher \\and Sebastian Altmeyer}
	\IEEEauthorblockA{
		Chair of Embedded Systems,
		University of Augsburg\\
		\{tilmann.laurenz.unte,johannes.kuehbacher\}@uni-a.de,\\ altmeyer@es-augsburg.de
	}
}

\maketitle

%
%

\input{abstract.tex}

\input{Section_Introduction.tex}

\input{Section_Related_Work.tex}

\input{Section_Usecase.tex}

\input{Section_RT_Cloud.tex}


\input{Section_Conclusion.tex}


\input{Suffix.tex}

%% file: comments.tex
\ifcomments
	\usepackage[colorinlistoftodos]{todonotes}
\else
	\usepackage[disable]{todonotes}
\fi



%% file: abstract.tex
%

\begin{abstract}

Edge computing processes data near its source, reducing latency and enhancing security compared to traditional cloud computing while providing its benefits. This paper explores edge computing for migrating an existing safety-critical robotics use case from an onboard dedicated hardware solution. We propose an edge robotics architecture based on Linux, Docker containers, Kubernetes, and a local wireless area network based on the TTWiFi protocol. Inspired by previous work on real-time cloud, we complement the architecture with a resource management and orchestration layer to help Linux manage, and Kubernetes orchestrate the system-wide shared resources (e.g., caches, memory bandwidth, and network). Our architecture aims to ensure the fault-tolerant and predictable execution of robotic applications (e.g., path planning) on the edge while upper-bounding the end-to-end latency and ensuring the best possible quality of service without jeopardizing safety and security.

\end{abstract}

\begin{IEEEkeywords}
	\textit{
		Safety-critical,
		Robotics,
		Cloud computing,
		Edge computing,
		End-to-end latency
	}
\end{IEEEkeywords}

%% file: Section_Introduction.tex
%

\section{Introduction}
\label{section:introduction}
Robots are part of emergent technologies used to perform complex physical tasks that are difficult or dangerous to perform by human labor alone. Clearly, safety is paramount in such applications. In the past, robots and human workers had to be physically separated to ensure safe operation. However, the current trend is to allow for collaboration without physical separation, leading to challenges related to the limited processing power and sensing capabilities of battery-powered robots, which are affected by SWaP (size, weight, and power) constraints. Increasing the sensor capabilities and processing power negatively influences the SWaP constraints, and it may be much harder to provide guarantees for real-time capabilities in such complex systems. On the other hand, more straightforward, low-power systems may not be able to fulfill the peak performance demands of their intended application.

As seen in recent projects such as SECREDAS\cite{gala-railway}, a new trend of deploying safety-critical systems on cloud computing platforms is emerging. Cloud computing can potentially revolutionize the field of robotics by providing a range of previously unavailable benefits. 
One of the critical advantages is the ability to offload heavy computation tasks to the cloud. Robots can use the cloud's processing power to offload and quickly complete resource-intensive tasks such as path planning and navigation algorithms or run artificial intelligence technologies to improve object recognition. This approach scales well and will allow for large numbers of robots to operate at the same time. Thus, battery-powered robots can specifically be optimized for their SWaP constraints and safety requirements. 
Robotics applications can be offloaded to the cloud as virtual machines (VMs) or using lightweight solutions such as containers or WebAssembly (WASM). Virtualization also helps to move away from custom hardware solutions by making the application layer independent of the underlying hardware and helps to solve hardware obsolescence issues. Cloud robotics will also enable robotics companies to enter a new market segment: \textit{robot operation as a service}.
Cloud computing can help combine data from multiple sources, including sensors, cameras, and other robots (including their sensor data), and process it in a timely manner, enabling robots to perform complex tasks and make more informed decisions. 
Moreover, by analyzing data from multiple robots, cloud-based machine learning algorithms can provide insights to improve performance and efficiency. 

One of the critical hurdles in realizing the full potential of cloud computing for robotics is that the cloud nodes can reside many (uncontrolled) network hops away from the robots' location, making it impossible to ensure high network bandwidth, low latency, bounded jitter, and thus, no end-to-end (E2E) timing guarantees as well. Edge computing is becoming increasingly popular as it enables data processing on edge nodes near the robot compared to the cloud. Thus, it helps minimize and upper-bound the worst-case network latency in transferring data to and from the cloud nodes. Low and bounded latency is crucial for real-time robotics applications. Additionally, edge computing offers better security as data is not transmitted over the internet, which reduces the risk of data breaches or cyber-attacks.

Multiple tasks (processes, VMs, containers, and WASM VMs) can run on each edge/cloud node and share the underlying node resources. Virtualization environments allow partitioning of CPU time and memory space for each task. Such pre-planned assumptions may be possible on single nodes, whereas, in edge computing with several nodes, each with several resources, these assumptions become less meaningful. Obtaining realistic resource availability assumptions becomes problematic if the applications, resource availability, or system configurations change. In current edge computing paradigms, no guarantee is given to tasks for many shared resources (e.g., memory bandwidth or network bandwidth). As a result, ensuring (timing) predictability or providing E2E guarantee to real-time safety-critical applications is an open issue demanding further research attention.

Previous works have looked into possible challenges, proposed theoretical architectures for using cloud/edge computing in the robotics domain (e.g., \cite{IRAAS2023, EdgeRobotics2022}), or presented initial results on offloading non-critical robotics applications \cite{SLAMoffloadingEnergy2023, EdgeRobotics2022}. Recent works such as \cite{VickRaaS2015} have proposed cloud computing service-based frameworks for motion planning applications to achieve hardware independence. Safety, security, and real-time capabilities have been recognized as critical requirements. Inspired from previous work \cite{DGRM,monacoISORC2023,gala-railway}, we propose an architecture for fault-tolerant and predictable execution of robotic applications (e.g., path planning) on the edge while upper-bounding the E2E latency and ensuring the best possible quality of service without jeopardizing safety and security.

The remainder of the paper is organized as follows: Section~\ref{section:usecase} introduces the industrial robotics use-case and discusses the advantages of using the edge cloud. Section~\ref{section:rtcloud} explains what parts of the robotics use case we can run in the edge and the requirements for enabling edge robotics. In Sections~\ref{section:components},~\ref{section:e2e}, and~\ref{section:ft}, we clarify how we intend to meet the requirements of the robotics applications in the edge.
Section~\ref{section:conclusion} concludes the paper and outlines the future work.

%% file: Section_Related_Work.tex
%
\section{Related Work}
\label{section:related_work}
\paragraph{Real-time cloud computing}
Abeni and Faggioli~\cite{abeni2019experimental} investigated the CPU latency of Xen and KVM hypervisors with and without PREEMPT\_RT patch for use with RT-cloud nodes. They provided some guidelines for correctly configuring the VMs to reduce the introduced latencies.
Several works, such as~\cite{Abeni2019ContainerbasedRS}, focused on (hierarchical) container scheduling to support real-time cloud computing.
\cite{Fiori2022RTkubernetesCR} and \cite{monacoISORC2023} presented Kubernetes orchestrator extensions to support the deployment of real-time applications (containers) in cloud infrastructure.
Gala et al.~\cite{gala-railway} evaluated existing cloud virtualization technologies for deploying a Real-Time (RT)-cloud to host an existing real-time safety-critical use case. They proposed a resource management layer to use with existing cloud virtualization technology (KVM) to support predictable safety-critical operation as a cloud-based service.
Abeni et al.~\cite{abeniFTRTC} presented the Fault-Tolerant Real-Time Cloud infrastructures capable of hosting highly reliable and real-time applications. 
Szalay et al.~\cite{RTFaaS} presented an architecture for a real-time Function-as-a-Service platform with the requirements for the underlying network and nodes.
The FORA project~\cite{barzegaran2023fora} addressed Industry 4.0 challenges from several angles, such as resource management and middleware, safety and security, and industrial control.


\paragraph{Cloud-based robotics}
There is ongoing research interest in combining industrial robots with cloud/edge computing infrastructure, with several surveys on the topic\cite{CloudRobotics2016, EdgeRobotics2022}.
Balogh et al. \cite{CloudBasedAMRs2021} proposed decoupling the closed-loop control of the robot from the robot’s embedded system and placing it into an edge cloud execution environment to benefit from ease of maintenance and improved resiliency to software/hardware failures while allowing the physical platform and control intelligence to evolve separately from each other.
Several works (e.g.,  \cite{SLAMoffloadingEnergy2023}) have looked into cloud/edge-based robot localization, mapping, or navigation.
However, the focus of these works is on performance measures or feasibility, not on the hard real-time constraints on the robot itself.
Vick et al. \cite{VickRaaS2015} proposed splitting the robotics applications into modular components capable of running on cloud servers while managing hard real-time tasks on the robot, leading to challenges that include guaranteeing reliable low-latency wireless communication\cite{4GCloudComputing, RTCloudRedundancy}. These challenges are in the scope of our proposed approach. Our aim is to dive deeper into the challenges regarding real-time constraints.


Our related work survey showed differences in computing power installed on the robot. In case of high computing power, the robot may be categorized as an edge device, allowing it to perform complex calculations~\cite{EdgeCloudPathPlanning}. In our work, we prioritize the hard real-time requirements and the SWaP constraints of the robot. This means it will use an energy-efficient rather than high-performance architecture, and tasks of lower criticality will be offloaded to a separate edge device.
In \cite{EdgeCloudPathPlanning}, the authors also utilize the cloud to facilitate collaborative robotics, which is out of scope in our approach. However, we leave it as an option for future work.
Extending the previous work on cloud/edge robotics, we tackle this topic with a main focus on the bounded E2E latency, security, and safety aspects.


%% file: Section_Usecase.tex
%

\section{Existing Industrial Robotics Use Case}
\label{section:usecase}
\outline{
	\begin{itemize}
		\item Detailed use case description
		\item Figure with existing robot components
		\item Flowchart / pseucode with most important parts - later we will use this and above figure to explain what will actually move to the cloud
	\end{itemize}
}
Our baseline is a case study from the realm of embedded real-time systems. It consists of a wheeled robot with sensors, a microcontroller with driver software for the hardware components, and a top-level application, which can include any number of extensions, e.g. additional sensors. 

\paragraph{Magni Robot}
\begin{figure}[t]
	\centering
	\includegraphics[width=0.7\linewidth]{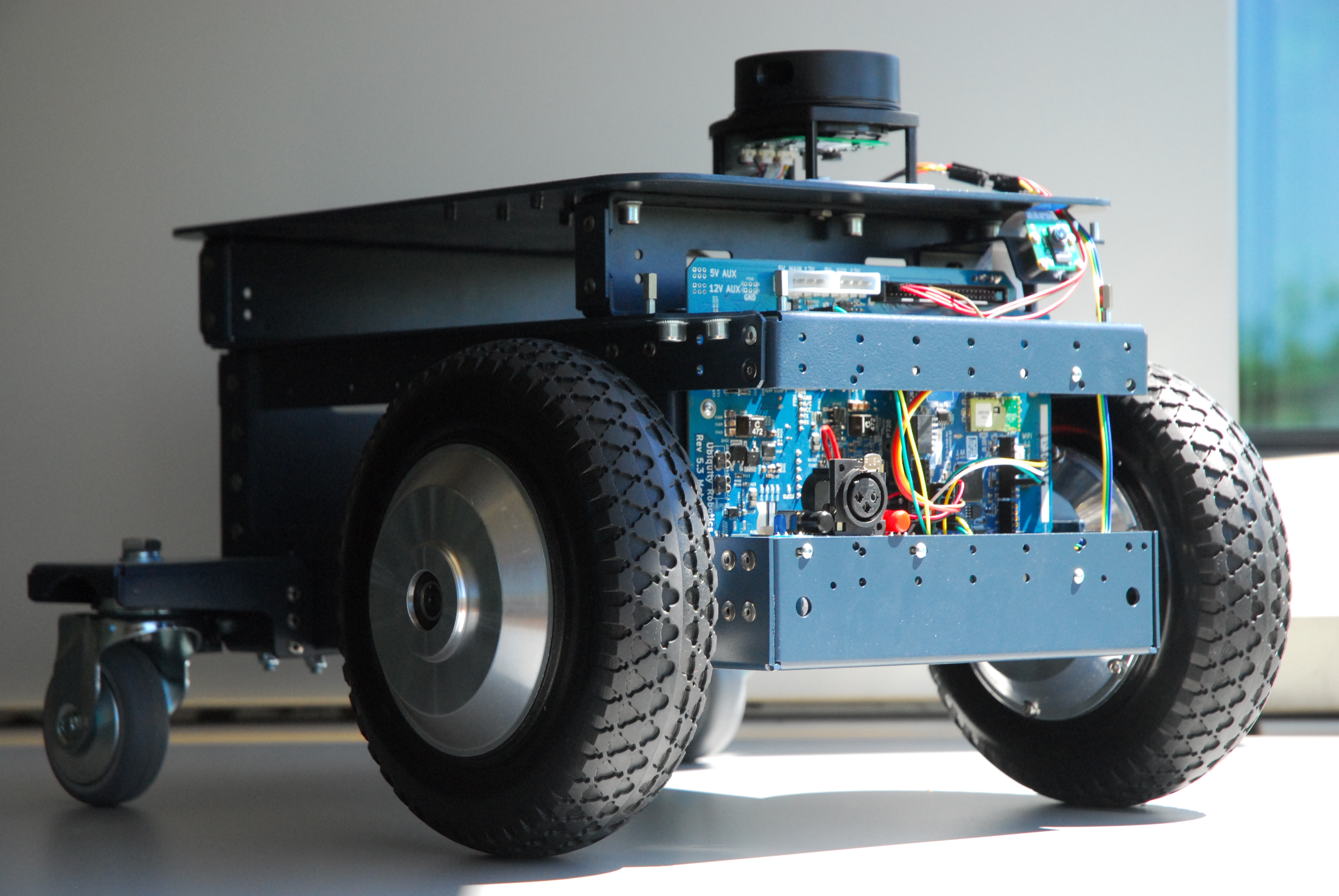}
	\caption{Ubiquity Robotics Magni robot with custom attachment for Slamtech RPLIDAR A1M8 sensor on its top plate}
	\label{fig:robot-pic}
\end{figure}

We looked for an industrial robot that offers us direct access to the hardware, not just a high-level software interface. We used the Ubiquity Robotics Magni, shown in figure~\ref{fig:robot-pic}. The Magni is a barebones system consisting of a \textit{Motor Control Board (MCB)} and two motors with internal rotary encoders. The robot's chassis allows for easy extension.
At the time of writing, the robot is shipped with a Raspberry Pi 4 running Ubuntu Linux and ROS 2. The manufacturer has released the ROS 2 software components communicating with the proprietary  MCB as open-source software. The communication takes place via a UART connection using a custom message protocol.

\paragraph{OS and Middleware}
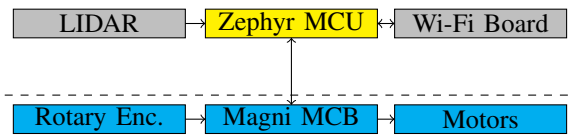
\begin{figure}[b]
	\centering
	\begin{tikzpicture}[scale=\linewidth*0.0150625]
		
		
		\draw [fill=lightgray] (0.03, 0.45) rectangle (0.63, 0.55);
		\node at (0.33, 0.5) {LIDAR};
		\draw [->] (0.63, 0.5) -- (0.7, 0.5);
		
		\draw [fill=lightgray] (1.36, 0.45) rectangle (1.96, 0.55);
		\node at (1.66, 0.5) {Wi-Fi Board};
		\draw [<->] (1.3, 0.5) -- (1.36, 0.5);
		
		
		\draw [fill=yellow] (0.7, 0.45) rectangle (1.3, 0.55);
		\node at (1, 0.5) {Zephyr MCU};
		\draw [<->] (1, 0.45) -- (1, 0.216);
		
		
		\draw [dashed] (0,0.25) -- (2,0.25);
		
		\draw [fill=cyan] (0.03, 0.116) rectangle (0.63, 0.216);
		\node at (0.33, 0.166) {Rotary Enc.};
		\draw [->] (0.63, 0.166) -- (0.7, 0.166);
		
		\draw [fill=cyan] (0.7, 0.116) rectangle (1.3, 0.216);
		\node at (1, 0.166) {Magni MCB};
		\draw [->] (1.3, 0.166) -- (1.36, 0.166);
		
		\draw [fill=cyan](1.36, 0.116) rectangle (1.96, 0.216);
		\node at (1.66, 0.166) {Motors};
	\end{tikzpicture}
	\captionsetup[table]{justification=raggedright,singlelinecheck=false}
	\caption{Our proposed hardware setup for a real-time capable robot controller. Below the dashed line are the proprietary components of the Magni robot. Above is our work, centered around a microcontroller running Zephyr OS.}
	\label{fig:robot-hw}
\end{figure}
We have redesigned this hardware and software stack to fulfill hard real-time guarantees. Our approach uses a single-core microcontroller instead of the Raspberry Pi. This way, we can facilitate tight WCET-bound analysis using state-of-the-art tools.
We chose to use the Zephyr real-time operating system for our software implementation. It offers a high degree of hardware abstraction, similar to the approaches used in the Linux Kernel. Our implementation extensively uses abstraction capabilities, splitting the project into middleware components, such as device drivers and hardware-independent top-level applications. Application programmers can pick and choose subcomponents as they need. The minimum viable configuration uses exactly one device driver for the Magni robot itself; however, additional sensor drivers are required for practical use.

In our implementation, the scheduler is configured for \textit{fixed-priority preemptive scheduling} with harmonic period tasks and implicit deadlines. Our proposed configuration uses five tasks listed in Table~\ref{tab:robot-taskset}. These tasks partially rely on the UART subsystem in Zephyr, which is hardware-dependent. The asynchronous nature of UART and the high transfer speeds mean hardware interrupts to handle it instead of software. 
\begin{table}[h]
	\centering
	\begin{tabular}{|c|c|c|c|}
		\hline
		\textbf{Task Name} & \textbf{Period} & \textbf{Priority} \\
		\hline
		Lidar Message Parser & \(1ms\) & \texttt{BASE\_PRIO} \\
		\hline
		Robot Message Parser & \(2ms\) & \texttt{BASE\_PRIO} \\
		\hline
		Robot Velocity Update & \(50ms\) & \texttt{BASE\_PRIO + 1} \\
		\hline
		Robot Odometry Update & \(250ms\) & \texttt{BASE\_PRIO + 2} \\
		\hline
		Robot System Manager & \(60000ms\) & \texttt{BASE\_PRIO + 2} \\
		\hline
	\end{tabular}
	\caption{A simplified overview of the base taskset used on the Zephyr MCU to facilitate communication between the robot, the Lidar, and the top-level application. The architecture-dependent UART interrupt subsystem used for the robot and the Lidar has been omitted.}
	\label{tab:robot-taskset}
\end{table}

\paragraph{Top-Level Application}
Our top-level application tackles the \textit{global localization problem}, i.e., we want to deduce the robot's pose within a known indoor environment. We use a Lidar to measure the distance to nearby obstacles and rotary encoders in the wheels to measure the robot's movement. The sensors are connected to the MCU using serial interfaces, as shown in Figure~\ref{fig:robot-hw}.

Our setup uses a Slamtech RPLIDAR A1M8 sensor as a custom extension to the Magni robot. This Lidar uses a single distance-measuring sensor connected to a motor that rotates continuously. For every rotation, the sensor takes \(360 \pm 6\) measurement samples. Each sample consists of two double-precision floating-point values corresponding to the rotation angle and distance. 
The application must perform measurement steps regularly to gather new sensor data. We assume that a full rotation of the Lidar is performed for every such step. Meanwhile, odometry data is gathered from the rotary encoders. These measurement steps can be bounded in terms of Bytes of data produced as:

$B_{Lidar} = (2 * 64) / 8$ and $B_{Odometry} = (4 * 64) / 8$
\begin{equation*}
	Total = (366 * B_{Lidar}) + B_{Odometry} = 5888 B < 6kB 
\end{equation*}
We assume that the robot stands still for Lidar measurements, which means the time taken for the measurement steps depends only on the Lidar, not the odometry sensors. 

\begin{equation*}
	ms_{Sample} = 0.5 ms
\end{equation*}
\begin{equation*}
	Total = (366 * ms_{Sample}) = 183 ms
\end{equation*}

Our localization application uses the \textit{Monte-Carlo Localization (MCL)} algorithm~\cite[p.~250]{thrun2005}:
The map is first filled with a large amount (e.g., \(10000\)) of random guesses of the robot's pose. Each guess is referred to as a particle. Together, the particles are used to approximate a probability distribution around the true pose of the robot, which the system can only ever approximate.
Iteratively, the robot is moved around the room, always trying to reach a goal position. With every movement command, all particles are moved accordingly. A small amount of noise is added to each particle to compensate for inconsistencies between the real robot and the movement model. Next, sensor data is gathered, which contains information about the relation of the robot's pose to the surroundings.
Based on a sensor model, the measurement step is repeated for all particles, and the results are compared against the real sensor values. This way, the fitness of each particle can be determined. A new set of particles is generated by randomly sampling from prior particles, where a high fitness yields a high likelihood of being selected. However, if the overall fitness is low, new particles may also be randomly generated. The population size never changes; however, as the confidence of the system increases, most particles will be similar.

Finally, the particles are combined to guess the robot's position. There are several possible approaches, such as using the weighted average of all particles or simply picking the one with the highest fitness. At this point, the path to the goal can be calculated, which we do using the A* algorithm~\cite{hart1968}. After a final check against the most recent sensor values to look for obstacles in the path, the robot will start driving towards the new goal, and the process repeats.

%% file: Section_RT_Cloud.tex
%
\section{RT-Cloud for Industrial Robotics Use Case}
\label{section:rtcloud}
\subsection{Why use edge-cloud?}
The particle filter demands high processing power. The main concern is the number of particles. 
In a complex or highly dynamic environment, 
one must assume the number of particles to be quite large, and in every step, every particle needs to be traversed by the algorithm.
Worse yet, the algorithm must calculate the sensor model's results for every particle. 

However, the algorithm can easily be parallelized, 
as particles have no data dependencies. 
Obviously, this is futile on a single-core microcontroller. 
In a first attempt to increase the processing power, 
we have separated our baseline into a trusted real-time component based around Zephyr and a single-core microcontroller together with a high-power component based on a Raspberry Pi 4. 
The on-board real-time system takes care of gathering sensor data and managing the robot itself, 
in particular when it comes to emergency halting. 
The Raspberry Pi receives the sensor data from the onboard system and runs the resource-intensive top-level application.

The main idea of our work is to use the edge's processing power to offload and complete resource-intensive top-level applications in a bounded time. Offloading to edge will eliminate the need for a Raspberry Pi or a more powerful processor onboard the robots. As a result, we can reduce the size and weight of the robots, thus decreasing production costs and making it easier to scale the number of robots. As evidenced in previous work \cite{SLAMoffloadingEnergy2023}, off-loading top-level applications (such as localization and mapping) to the edge may also help reduce the robots' power consumption and increase battery life. 
Moreover, the resources at edge nodes can be utilized to improve the availability and re-usability of top-level applications. It will also make it easy to reconfigure and replace the top-level application without physical access to the robot. Overall, we envisage high scalability, less maintenance, and reduced costs.


\subsection{What can we run in the edge?}
\outline{
	\begin{itemize}
		\item Entire Monte-Carlo Localization Algorithm
		\begin{itemize}
			\item Sensor Data Processing (potentially in parallel?)
			\item Sensor Model (for each particle, in parallel)
			\item Particle Resampling (probably strictly sequential)
			\item Movement Model (for each particle, in parallel)
		\end{itemize}
	\end{itemize}
}

\begin{figure*}
	\centering
	\includegraphics[width=\linewidth]{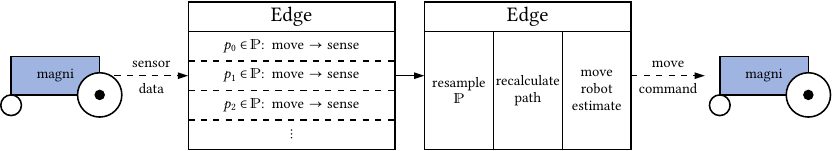}
	\caption{Outline of the individual steps of the Monte-Carlo Localization (MCL) algorithm running on an edge device. The dashed arrows indicate wireless communication. The Magni diagrams shown on the left and right correspond to a single physical robot and can be seen as the start and end points of the complete control loop.}
	\label{fig:mcl-steps}
\end{figure*}

In Figure~\ref{fig:mcl-steps}, we show the high-level vision of the MCL algorithm offloaded to the edge device.
The robot gathers sensor data as usual, but instead of processing it directly, it transmits it wirelessly to the edge device. Now, the edge device computes the MCL. The first step is the movement model and sensor model in the particle filter. Here, we can make the most of the edge's additional processing power, as this part of the algorithm allows for massive parallelism.
Once all particles have been updated, the algorithm reaches a strictly sequential section, where particles are resampled according to their fitness. Finally, a new path to the desired goal is calculated, which can then be translated into a move command. This move command is returned wirelessly to the robot.

\subsection{What are the requirements for running in the edge?}\label{sec:requirements}
Robotics applications require dependable and predictable operation with bounded E2E latency. These requirements translate into dealing with faults (e.g., untimely arrival or omission of communicated data), encapsulating applications for safety and security, and providing E2E guarantees.

\begin{enumerate}
	\item The robotics application running on the edge require encapsulation for safety and security. Errors, faults or security issues in other applications should not trigger
	any failure on the robot application or compromise the node itself. A way to achieve this is by adequately partitioning the
	different resources used by the application.
	\item Multiple robotics applications running on the same edge hardware require predictable timing behavior (i.e., predictability despite multicore interference). Applications require some fixed minimum allocation of resources. For example, the virtualization layer must assign each application the minimum required time on the CPU, and a fixed predefined amount of cache and memory allocation, and a fixed memory and network bandwidth.
	\item Robotics applications running on the edge must be able to connect safely to the robot. This involves predictable timing and orderly message delivery between different apps or among apps and robots. Apps require a minimum guaranteed network bandwidth to achieve this.
	\item The virtualization overhead must be negligible as compared to running the application natively on the robot itself.
	\item Faults can affect predictability and must be dealt with, e.g., a drop of packages, traced back to communication conditions, must be minimized. 
	\item Monitoring is required to detect faults or anomalous behavior of applications and availability of resources.
	\item Predictable timing behavior for re-orchestration of applications upon occurrence of faults/failures.
\end{enumerate}

Adequate security measures and tooling are required to fend off threats and maintain safe and secure system operations.
External influences of accidental or malicious origins must not compromise the integrity or authenticity of communicated data.
Dealing with cyberattacks is essential 
as they can exploit unpatched vulnerabilities in the system in order to compromise edge nodes or the shared network and trigger system failures. 
\begin{enumerate}
	\item Security approach must be generically applicable and independent of the application type.
	
	\item The Security approach must be invisible to the applications.
	
	\item For safety protection, robotics applications running on the edge must communicate securely over private networks, and their communicated data remain integral and authentic.
	
	\item Malicious attackers must not be able to break through deployed defenses for communication or at nodes.
	
	\item Enforcing security must not come at the cost of meeting the E2E timing requirements of the applications.
\end{enumerate}


\subsection{Edge component selection}
\label{section:components}

We made several architectural decisions to enable the robotics application to run in a virtual environment on the edge node.
Figure~\ref{fig:edge-node} depicts the architecture of such an environment.
We can identify some novel major components compared to the traditional approach of running the robotics applications on dedicated hardware: 
\begin{figure*}
	\centering
	\includegraphics[width=\linewidth]{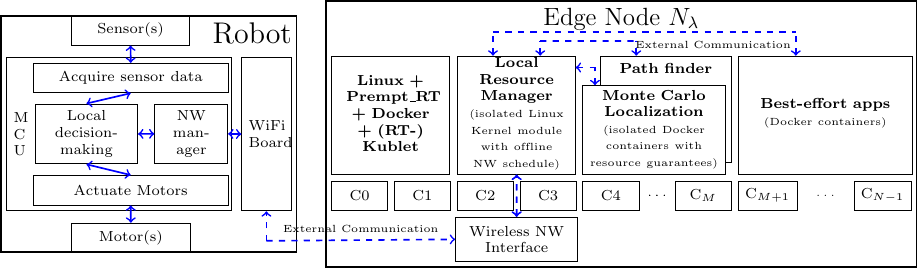}
	\caption{Robot and Edge Node Architecture}
	\label{fig:edge-node}
\end{figure*}

\paragraph{Virtualization layer}
Containers have become the de facto industrial approach, especially in cloud environments, providing an ideal underlying layer for edge-to-cloud and multi-cloud scenarios. Hence, we use Linux with Docker containers to separate, isolate, and abstract the safety-critical applications from the COTS hardware.
To fully take advantage of the edge node resource and virtualization, we must run multiple robotics and best-effort application containers on the same edge node. However, multiple containers on the same edge node share the underlying resource, such as the network interconnection, last-level cache (LLC), and memory bandwidth. Such resource sharing can cause unpredictable delays and, thus, unbounded E2E latency for robotics applications.
Linux and Docker alone are not adequate to meet the requirements for running the robotics application on the edge node and provide bounded E2E latency.


\paragraph{Fine-tuning of edge nodes}
Let us assume a multicore edge node with $N$ CPU cores ($C_0,C_1,\ldots,C_{N-1}$) running Linux (and Docker).
Out of these $N$ cores, we dedicate cores $C_{2},C_{3},\ldots,C_{M}$ ($M<N$) to run critical applications. We will refer to these cores as critical cores.
We can use Cores $C0, C1$ and $C_{M+1},\ldots,C_{N-1}$ ($M<N$) to run Linux (+Docker) and best-effort docker containers, respectively. We will refer to these cores as best-effort cores.
To obtain real-time behavior from the Linux kernel (low latency and high determinism), we chose the fully preemptible kernel (RT) model via kernel configuration during build time. 
The standard Linux kernel performs significant asynchronous housekeeping work, such as timekeeping, timer callbacks and interrupt handlers, on all cores. The ``noise'' from such work can significantly impact the predictable execution of robotic application containers.

Linux \textit{isolcpus} parameter helps us to isolate the cores $C_{2}, C_{3},\ldots, C_{M}$ from the general SMP balancing and scheduler algorithms.
Similarly, we initialize the $nohz\_full$ to configure $full~dynticks$ along with CPU Isolation (assuming only one task per CPU core), $rcu\_nocb\_poll$ to offload RCU processing to the best effort cores, and the $irqaffinity$ parameter to affine the IRQs to best effort cores (e.g., $nohz\_full=C_{N-M}, C_{N-M+1},\ldots, C_{N-1}$).
Once the system boots, Linux ensures that no processes execute on the critical cores unless instructed and restricts all housekeeping work (including those from docker) to the best-effort cores. Thus, we ensure almost housekeeping noise-free critical cores to run the robot applications.

\begin{figure}[t]
	\centering
	\begin{subfigure}{0.5\textwidth}
		\centering
		\includegraphics[width=\linewidth]{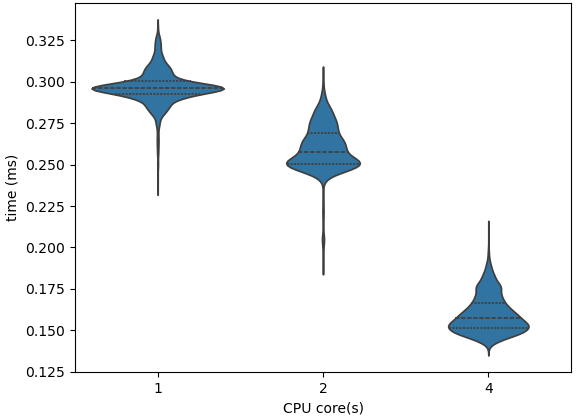}
		\centering
		\caption{In isolation with increasing number of CPU cores}
		\label{fig:MCL-CPU1-4}
	\end{subfigure}
	\begin{subfigure}{0.5\textwidth}
		\centering
		\includegraphics[width=\linewidth]{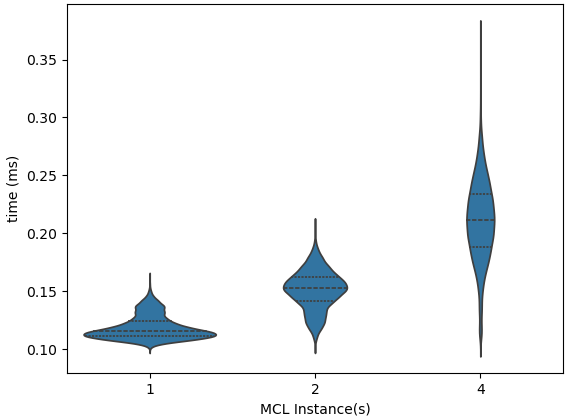}
		\centering
		\caption{With multiple instances of MCL containers in parallel}
		\label{fig:MCL-Stress}
	\end{subfigure} 
	\caption{Observed execution time for MCL + Path Finder}
\end{figure}

We ran the Monte Carlo Localization and Path Finder docker containers in isolation on critical cores $C_{4}, C_{5},\ldots, C_{M}$. Figure \ref{fig:MCL-CPU1-4} shows the total observed execution time over 50 runs of the MCL and Path Finder applications with an increasing amount of CPU core allocation (1 to 4 CPU cores).
We also ran multiple instances (up to 4) of Monte Carlo Localization docker containers in parallel to simulate a use case with multiple robots. Figure \ref{fig:MCL-Stress} shows the total observed execution time over 50 runs. The results demonstrate that Linux and Docker alone cannot meet the requirements for running the robotics application on the edge nodes and provide bounded E2E latency. We need additional resource management mechanisms to ensure such requirements, especially due to shared resource contention on multicore nodes.

\paragraph{Container orchestration}

Container orchestrators, such as Kubernetes (K8s) \cite{k8s}, manage, deploy, and scale containers (e.g., Docker) across multiple nodes. Orchestrators allow easy deployment of containers by properly setting up their runtime and monitoring their execution without the need for a cumbersome and error-prone manual process. Usually, orchestrators manage only a few fixed resources, such as CPU time and memory space, and deliver acceptable Quality-of-Service (QoS) to various users. Orchestrators deploy containers based on the user requirements while guaranteeing and balancing the usage of these few resources across the nodes. Moreover, we can configure orchestrators to deploy container replicas on different nodes upon failures. Orchestrators can help exploit resource isolation possibilities provided by the underlying hardware and software layers to support meeting RT requirements.

Widely used container orchestrators (e.g., K8s), the underlying Linux kernel, container technology (e.g., Docker), and built-in resource management approaches (e.g., Linux Cgroups) are geared towards average-case performance and not designed to consider strong shared resource isolation and E2E guarantees. Moreover, in a system with dynamically changing availability and demand for resources, the orchestrator must be aware of all containers and the availability of shared resources on each node (e.g., memory bandwidth, cache, and shared interconnect). The orchestrators must coordinate edge-wide resources and adapt the containers to the current availability of resources (and not simply suspend the best-effort containers). 

Currently, K8s only supports the orchestration of CPU (temporally) and memory (spatially), not other shared resources (e.g., memory bandwidth). 
In addition, we also need to employ some form of network virtualization and separate the traffic from different containers on several edge nodes while still predictably transporting it over the same shared physical medium.

\paragraph{Resource management and orchestration (RMO)}
The new RMO layer and Linux and Kubernetes extensions are inspired by previous works such as \cite{DGRM, monacoISORC2023}. 
As suggested in \cite{monacoISORC2023}, we envisage a RMO layer for the edge that can be integrated with Linux, Docker, and Kubernetes orchestrator. 
In our RMO layer, each node contains a node-level Local Resource Manager (LRM) to manage its resources in coordination with Linux and Kubernetes.
The LRM is comprised of:

\begin{enumerate}
	\item Monitors (MON) to check the availability of resources on a node and the behavior of critical and best-effort containers.
	They can be hardware units (e.g., performance monitor counters), monitors built into the hypervisor/OS, or monitors specially developed for robotics applications.
	
	\item Local Resource Schedulers (LRSs) to perform run-time scheduling of resources.
	LRSs can be hardware units (e.g., Intel Xeon's memory bandwidth allocation~\cite{xeon}), schedulers integrated into operating systems and hypervisors (e.g., Linux's Sched\_Deadline), or specially developed schedulers to ensure predictable access to resources.
\end{enumerate}


The Local Resource Manager (LRM) manages all node's resources based on the monitoring information provided by the Monitors (MON). 
The LRM assigns the required resources to the containers via LRSs to ensure that the robotics application has bounded E2E latency and meets their safety assurance levels while the best-effort containers can get the best possible QoS.
Because of resource failures or changes in conditions at run-time, if node-level scheduling of a container is not possible, 
the LRM coordinates with the LRMs of other edge nodes to find nodes that can accommodate these containers.
We can achieve coordination and communication among the LRMs via Kubernetes with real-time extensions \cite{monacoISORC2023}.



\paragraph{Wireless network}
\outline{
	\begin{itemize}
		\item IEEE 802.11
		\begin{itemize}
			\item How it works and limitations
			\item How to provide a timeliness and reliable wireless network 
		\end{itemize}
		\item Assumptions and Requirements
		\begin{itemize}
			\item IEEE 802.11
			\item All nodes in the NW are visible by all the other nodes
			\item Decentralized network
			\item Access to the communication medium according to a TDMA schedule
			\item Node synchronization 
		\end{itemize}
	\end{itemize}
}
IEEE 802.11 is the standard that defines the set of protocols necessary for implementing a wireless local area network (WLAN). 
In a WLAN, nodes communicate with each other over a shared communication medium, where interference and overlapping transmissions can occur, causing frames to have unbounded E2E latencies.
In the IEEE 802.11 standard, DCF (Distributed Coordination Function) is the primary method for coordinating node access to the communication medium. 
DCF employs CSMA-CA (Carrier Sense Multiple Access with Collision Avoidance), which means that a frame transmission only starts after DCF \textit{senses} that the communication medium is \textit{idle}. 
In order to avoid transmission interference, consecutive frames from a node are spaced by bounded delays known as IFS (Interframe Space delays).
However, for a time-sensitive application like ours, the collision avoidance capabilities of DCF and the IFS delays are not enough to ensure bounded transmission time for the frames sent (received) by the robot to (from) the cloud.

A protocol that provides timeliness and reliable wireless communication is necessary to obtain bounded E2E latencies in a WLAN. 
For our edge-robot application, we implement a WLAN based on the TTWiFi protocol proposed by Lusty et al. \cite{lusty2021ttwifi}.  
TTWiFi is a time-triggered (TT) communication protocol merged with the IEEE 802.11 standard providing transmission time upper bounds for all the nodes connected to the network.

\paragraph*{Network Details}
In our implementation, we consider the IEEE 802.11n extension due to its extensive presence in wireless devices, as well as its range and data rate capabilities.
Concerning the network topology, we assume that all nodes in the network are \textit{visible} to each other. 
If a node is within the transmission range of another node, we say that those two nodes are visible to each other.
As a result, we eliminate the hidden node problem, which could lead to data replication and additional delays due to frames being re-transmitted.
We use a decentralized wireless network instead of a centralized access point (AP) to avoid a single point of failure and increase the reliability of our WLAN. 
We assume that each node in the network can only access the communication medium at specific points in time according to a cyclic TDMA (Time Division Multiple Access) schedule. The LRM of the RMO layer can enforce such a network schedule.
We consider that the WLAN has a clock synchronization mechanism that allows all the network nodes to adjust their own global clock based on timestamps exchanged by the nodes during transmissions.


\section{End-to-End Guarantees}
\label{section:e2e}
\outline{
	\begin{itemize}
		\item Communication Coordination
		\begin{itemize}
			\item Synchronizationpotentiality
			\item Transmission Scheduling
			\begin{itemize}
				\item TDMA
			\end{itemize}
		\end{itemize}
		\item Transmission Time
		\begin{itemize}
			\item Source of delays
			\begin{itemize}
				\item Low level and high level
			\end{itemize}
		\end{itemize}
		\item MAC level modifications
		\begin{itemize}
			\item Disable CS in DCF
			\item MAC level modification
		\end{itemize}
		\item Bounded Transmission Time
		\begin{itemize}
			\item Transmission Slot
			\item Interframe Space
			\item Bit-rate
		\end{itemize}
		\item Problems to Avoid
		\begin{itemize}
			\item Hiden Nodes
			\item Scheduling Jitter
			\item Packet Fragmentation
			\item Network scanning
		\end{itemize} 
	\end{itemize}
}
In order to achieve bounded E2E latencies in a WLAN, it is essential that all the possible 
sources of unbounded delays have been removed from the network or bounded. 
In the following subsections, we describe which mechanisms need to be added to our WLAN and which modifications in the IEEE 802.11n protocol are needed to upper-bound the E2E latencies. 

\paragraph{Bounded Transmission Time}
In a standard WLAN, different delays contribute to non-deterministic transmission times.
The amount of delay suffered by a frame before and after its transmission depends on which level of the WLAN those delays originate.
Like Lusty et al. \cite{lusty2021ttwifi}, we classify possible sources of delays into two categories: (I) High-level delays and (II) Low-level delays. 
High-level delays originate outside the IEEE 802.11 standard, e.g., delays from the transport layer of the network. 
Low-level delays originate within the IEEE802.11 standard, e.g., the backoff timer in DCF. 

We minimize high-level delays by using UDP instead of TCP as our communication protocol. 
UDP provides a minimalist implementation that results in low temporal overheads and jitter compared to TCP. 
By not using TCP, we avoid transmission throttling and packet re-transmissions, which are inherent properties of TCP that affect the amount of delay a packet might suffer during transmission. 
Unlike TCP, UDP does not prevents packets from being duplicated.  
However, since we consider a decentralized wireless network instead of a centralized AP, packet duplication is not a problem in our network as frames are not re-transmitted. 
Although UDP is not free of temporal overheads, e.g., delays caused by the error check mechanism, those overheads are low and predictable.

We minimize low-level delays by disabling the CS (Carrier Sense) mechanism of DCF, which is used to identify if the communication medium is idle or not.
If it is busy, the node attempting to access the medium backs off for an unbound time. 
The backoff time is one of the primary sources of unbounded delays in a WLAN since it depends on internal and external factors of the WLAN.  
Since in our network nodes only access the communication medium at predefined time points according to a TDMA schedule, there is no need to check whether the communication medium is busy.  
Therefore, the CS mechanism becomes irrelevant and can be safely disabled. 

Network scanning and packet fragmentation are other sources of delay at the low level present in a standard WLAN. 
As the name suggest, packet fragmentation fragments packets into smaller parts in order to improve reliability and efficiency during transmission. 
However, packet fragmentation results in a nondeterministic temporal behavior due to the additional invocations of the DCF mechanism. 
Since we consider UDP as our primary communication protocol, we mitigate the packet fragmentation problem as the IEEE 802.11 standard does not allow packets being broadcast to be fragmented. 
In a standard WLAN, it is common for nodes to periodically pause their transmissions and change their transmission frequency. 
They do that to scan the network in search for other nodes that might be part of the network. 
However, this \textit{scanning} behavior leads to additional delays that impact the frame transmission time.  
Since we consider that the number of nodes in the network is fixed and that all nodes are visible to each other, we disable node's network scanning capability.

\paragraph{Node Synchronization}
\label{sec-nodeSync}
A common way to synchronize nodes in a network is by exchanging timestamps. 
By doing so, each node receives a timestamp and estimates the sending node's clock based on an approximated transmission time and the received timestamp.   
In our edge-robot application, transmissions might overlap if nodes are not correctly synchronized. 
By recording the actual arrival time of a frame and comparing it with the expected arrival time, a node can estimate its clock difference from other nodes. 
However, synchronizing clocks using timestamps may not correctly represent when a transmission has started.   
For example, a frame might not be transmitted immediately after it has been timestamped because the OS had to serve a higher priority process. 

In order to overcome this issue, we assume that a transmission process cannot be preempted right after it timestamps a frame. 
On top of that, we add a guard time between TDMA slots to account for other possible delays caused by the OS.  
During guard time, no node in the network is allowed to transmit. 
This extra time helps to cover the possible error margin existent due to the synchronization mechanism.    
Therefore, by periodically recomputing nodes' global clock based on scheduled transmissions and timestamps sent by other nodes, we ensure that all nodes in the network are synchronized.  
To reduce the effect of outlier clock values, we consider that the re-computation of global clock is based on the Fault Tolerant Average protocol \cite{kopetz1987clock}.

\paragraph{Transmission Scheduling}
Since we consider fixed-priority preemptive scheduling, we elevate the priority of the transmission process to the highest possible level in the system to minimize scheduling jitter. 
Therefore, we rely on the RMO layer (edge node) and the real-time capabilities of the Zephyr board (robot) to ensure that nodes schedule and start their transmissions within the assigned TDMA slot.

\paragraph{Implementation Aspects}
In order to achieve bounded transmission time in our WLAN, we have to make a few modifications to the MAC behavior of the IEEE 802.11 standard. 
Like Lusty et al. \cite{lusty2021ttwifi}, we use in our edge-robot application the ModWifi project \cite{mod2015}, which allows users to reconfigure some parameters of the IEEE 802.11 standard, e.g., the CS mechanism behavior, IFS values, and backoff queue lengths. 
The ModWifi project is built on top of the open-source Qualcomm Atheros 802.11n driver and firmware for the AR7010 and AR9271 network chips. 




The most important modification we do is to \textit{disable} the CS mechanism present in DCF since leaving it enabled would lead to unbounded temporal delays.
We \textit{disable} the CS mechanism by setting to true the \textit{force channel idle} parameter in ModWifi. 
We also set the IFS values (SIFS, EIFS, and AIFS) to the minimum value possible, e.g., 1$\mu$s.
By doing those modifications and setting to 0 the \textit{aSlotTime} parameter (back off time), we allow predefined transmissions to start more accurately in accordance with the TDMA schedule. 
According to our estimates, the 5888 Bytes data produced by the robot's sensors (see Section~\ref{section:usecase}) need three 802.11 frames. 

\begin{equation*}
	\frac{frameSize}{bitRate} + IFS
\end{equation*}

For e.g., considering a bit rate of $400Mbit/s$ and a IFS of $1\mu s$, the upper bound to transmit robot's sensor data is $122.16\mu s$. 
 
\section{Fault-tolerance and Security Aspects}
\label{section:ft}
\outline{
	\begin{itemize}
		\item faults in edge
		\begin{itemize}
			\item we assume hypervisors to be free of design faults and do not behave maliciously.
			\item Container running MCL for given sets of particles may behave arbitrarly, but their faults can be masked by active replication with voting (patching FT2) 
		\end{itemize}
		\item faults in robot
		\begin{itemize}
			\item we assume Zephyr board only to suffer crash faults causing the robot to stop, justified by simplicity of the board and the application it's running.
		\end{itemize}
		\item faults in communication channel
		\begin{itemize}
			\item communication loss with edge: use-case belongs to fail-safe systems, Zephyr board must not issue drive or rotation commands until communication re-established
			\item we can attempt to improve reliability of message delivery by sending multiple replica messages and over independent communication channels (wireless bands) 
			\item Hypervisors and robots (Zephyr boards) do not exhibit malicious behavior and hence will follow the TDMA schedule. They do not send outside their TDMA slots. (patching FT1)
		\end{itemize}
	\end{itemize}
}

\paragraph{Edge Nodes}
We assume that edge nodes follow a hybrid fault model, 
where containers (responsible for computing the localization information) can fail arbitrarily and maliciously,
whereas their underlying virtualization layer may only exhibit crash faults, silencing the whole edge node in the process.
Such a hybrid fault model is common~\cite{syncguard2023} and can be justified by the strong isolation of the virtualization layer. 
We envisage replicating the offloaded localization algorithm and deploying it over multiple containers on several edge nodes.


We plan to fortify our system against persistent cyberattacks that can successfully compromise the system over time by exhausting the majority of containers in replica groups. 
This can be achieved by having the RMO layer coordinate the rejuvenation of containers such that any adversarial presence is removed~\cite{proactiveReactiveRecovery2010}. 
Furthermore, eliminating adversarial knowledge from previous infiltration attempts requires implementing techniques such as obfuscation~\cite{proactiveObfuscation2010} and address space randomization~\cite{runtimeRandomization2003}. 

\paragraph{Robot}
The simplicity of the functionality implemented on the robot's Zephyr board justifies the adoption of a fail-silent mode. 
i.e., malicious and Byzantine behaviors are precluded.  
The board implements rudimentary and timely self-checking mechanisms. 
Upon failure detection, the robot halts until a safe state is reached or maintenance is performed.  

\paragraph{Communication}
We assume that the local shared network is protected against external and intentional interference (e.g., shielding the environment where the application is deployed). 
Jamming devices planted within the environment are outside our fault hypothesis since no known solution for this problem exists under wireless-based networks.   
Therefore, we focus on ruling out interference caused internally by nodes on the network. 
We plan to achieve this by having the RMO layer and Linux enforce compliance with the TT communication schedule.  
Hence, no compromised container will successfully transmit outside its allocated slot, avoiding interference and dropped packets. 
The Zephyr board uses a fail-silent model and is trusted to follow the communication schedule. 
Otherwise, additional trusted components can be used to control network access in the time domain \cite{syncguard2023}. 

The reliability of message delivery over the wireless network can be improved by replicating messages in the time domain, 
i.e., transmitting multiple instances of the message in the same TDMA slot. 
Reliability can also be improved via space domain replication,
e.g., the same message can be sent over multiple independent networks (using different communication bands). 
However, this approach is costlier as it requires more communication hardware. 

Encrypting data for confidentiality incurs heavy computational overheads, 
which potentially affects the ability to meet the timing requirements. 
Moreover, confidentiality is currently not a concern for the application.
We focus on protecting the integrity and authenticity of information communicated between edge nodes and the robot by providing a secure communication service. 
An authentication protocol based on time-delayed release of keys is a good candidate for our application \cite{TTsec2011}. 
Its procedures are lightweight and designed specifically to work with TT communication.  
The secure communication service shall fend off attempts to trick the robot or edge nodes into accepting spoofed or faked messages.

%% file: Section_Conclusion.tex
%


\section{Conclusion}
\label{section:conclusion}
\outline{
}

This paper advocates for the strategic offloading of parts of an existing safety-critical robotics use case from an onboard dedicated hardware solution to the edge. This transition to the edge harnesses its high processing power, leading to a significant reduction in the robot's size, weight, and power requirements. The edge's low latency, a crucial factor for this use case, is complemented by the cloud computing principles it offers, supporting scalability, reusability, high availability, and hardware independence (via virtualization).
The use case requires fault-tolerant and predictable execution of robotic applications on the edge while upper-bounding the end-to-end latency and ensuring the best possible quality of service without jeopardizing safety and security.
We proposed an edge architecture based on Linux, Docker containers, Kubernetes, and a TTWiFi local wireless area network to offload the robotics applications. Based on insights from previous work on real-time cloud, we added a resource management and orchestration layer to ensure we meet the use requirements.

Currently, we have all the individual components for the edge robotics use case: a Magni robot with a Lidar sensor and Zephyr RTOS, Kubernetes edge nodes with resource management and orchestration enhancements, a Monte Carlo Localization and Path Finder applications running in docker containers, and a wireless network with the TTWiFi protocol. We ran some initial experiments to observe the worst-case execution time of Monte Carlo Localization and Path Finder applications containers on the edge node and calculated estimates for gathering the sensor data from the robot and transmitting the data to the edge nodes. We also provided suggestions for ensuring fault tolerance and security. The next step is putting all these individual steps and components together and demonstrating predictable execution and upper-bounded end-to-end latency of robotic applications on the edge.

%% file: Suffix.tex
%

\bibliographystyle{IEEEtran}      
\bibliography{References}   

\end{document}